\documentclass[sigconf]{acmart}

\usepackage{listings} 
\usepackage{tikz}
\usetikzlibrary{shapes, arrows.meta, positioning}
\usepackage{hyperref}
\usepackage{microtype}
\usepackage{booktabs, caption}
\usepackage[colorinlistoftodos]{todonotes}
\usepackage{threeparttable}
\usepackage{xcolor}

\AtBeginDocument{%
  }

 \setcopyright{cc}

\acmBooktitle{Preprint}

\begin{document}

\lstset{
    backgroundcolor = \color{white},
    numbers=left,
    numberstyle=\tiny,
    numbersep=5pt,
    xleftmargin=11pt,
    xrightmargin=4pt,
    frame=single,
    aboveskip=0pt,
    belowskip=-6pt,
    sensitive=true,
    float=!t,
    breaklines=false,
    captionpos=b,
    tabsize=2,
    showstringspaces=false,
    basicstyle=\small\ttfamily,
    morecomment=[l]{//},
    morecomment=[s][\itshape]{/**}{*/}
}

\title{Using Grammar Masking to Ensure Syntactic Validity in LLM-based Modeling Tasks}

\author{Lukas Netz}
\email{netz@se.rwth-aachen.de}
\orcid{0000-0003-2013-2919}
\affiliation{%
  \institution{Chair of Software Engineering}
  \city{Aachen}
  \state{NRW}
  \country{Germany}
}

\author{Jan Reimer}
\email{jan.reimer@rwth-aachen.de}
\affiliation{%
  \institution{Chair of Software Engineering}
  \city{Aachen}
  \state{NRW}
  \country{Germany}
}

\author{Bernhard Rumpe}
\email{rumpe@se.rwth-aachen.de}
\orcid{0000-0002-2147-1966}
\affiliation{%
  \institution{Chair of Software Engineering}
  \city{Aachen}
  \state{NRW}
  \country{Germany}
}

\renewcommand{\shortauthors}{Netz et al.}

\begin{abstract}
        %
        We present and evaluate a method called grammar masking, which is used to guide large language models (LLMs) toward producing syntactically correct models for a given context-free grammar.
        %
        %
        Prompt engineering methods such as few-shot learning or priming can be used to improve the chances of an LLM producing correct syntax, but the more complex the grammar, the more time-consuming and less promising these methods become.
        %
        %
        Previous work is focused primarily on the usage of either language model training or prompt engineering.
        %
        %
        In this work, a method is presented that restricts the output to a given grammar using constrained decoding to ensure the output adheres to a valid syntax.
        %
        %
        We use several DSLs built with MontiCore and task multiple LLMs to produce models with and without constrained decoding. A corresponding parser is used to confirm the syntactic correctness of each model.
        %
        %
        We show that grammar masking can dramatically improve the modeling capabilities of several LLMs, reducing the need for well-refined prompting while increasing the chance of producing correct models.
\end{abstract}



\maketitle

\section{Introduction}

Large language models (LLM) \cite{liu2023summary,zhao2023surveylargelanguagemodels} are highly sophisticated tools that, among other things, are capable of generating code artifacts based on a natural language input \cite{barenkamp2020applications, crawford2023ai, sadik2023coding}. As we operate in the context of model-driven software engineering, we focus on the synthesis of textual models using the predefined syntax of a given domain-specific language (DSL). Given that the syntax definition of the targeted DSL might not be included in the corpus of training data used for the language model, it is necessary to rely on post-training optimization techniques such as few-shot learning \cite{brown2020language, fedele2023explain}, fine-tuning \cite{ren2018meta}, or prompt engineering \cite{chen2024unleashingpotentialpromptengineering, muktadir2023brief}. However, as these methods rely on prompt engineering, they have one common element: they only improve the likelihood that the LLM produces syntactically correct models but cannot guarantee it. 

In this work, we introduce an approach that uses the context-free grammar (CFG) of the targeted DSL to filter out any syntactically invalid output during the generation process of the LLM. We will evaluate results by comparing this approach to previous successful modeling tasks for LLMs using only few-shot learning.

\section{Foundations}
We introduce several foundations, such as the used framework Guidance, and the DSLs for which we will generate models.

\subsection{Large Language Model}

A LLM is a language model that is trained on a vast amount of text data. It is distinguished by its capability for general-purpose language understanding and generation. These models acquire their abilities by learning statistical relationships from text documents through a computationally intensive self-supervised and semi-supervised training process \cite{Vaswani+17}. LLMs can perform text generation, a type of generative AI \cite{amaratungaunderstanding}, by taking an input text and iteratively predicting the next token or word.
In the context of software engineering, LLMs present significant potential to enhance and automate various tasks \cite{macneil2023experiences}, particularly those related to Model-Driven Software Engineering (MDSE) and modeling languages \cite{baumann2024combining, netz2024natural}.

\subsection{Few-Shot learning}
Few-shot learning (FSL) is a well-established in-context learning approach for large language models \cite{brown2020language, fedele2023explain, ren2018meta}. A pre-trained LLM can be prompted with a set of $N$ exemplary question-answer pairs $(q^{i}, a^i)^N_{i=1}$ before being provided with the actual question $q$. The FSL output $a$ for the question $q$ is defined as $P_{LLM}(a | q, (q^i,a^i)^N_{i=1})$. In addition, further instructions can be added to improve the results \cite{wei2021finetuned}. Further work is published on the FSL improvement introducing intermediate reasoning steps (e.g. Chain of thought) \cite{wei2022chain, dohan2022language, wang2022self}.

The success of few-shot in-context learning depends on the utility of the implicit knowledge within the provided examples and the clarity with which the task specifications are communicated through the provided examples. In the case of Domain-Specific Languages, the structured nature of the combinatorial output space, represented by the CFG of the DSL, is not easily covered by the limited number of demonstrations. Thus, generating models for a DSL with an FSL-based approach remains a significant challenge for LLMs.

\subsection{Guidance-AI}
\label{sec:guidance}

\begin{figure}[h]
    \centering
    \begin{tikzpicture}[
        level distance=1.5cm,
        sibling distance=2.5cm,
        edge from parent/.style={draw, -{Latex}},
        every node/.style={font=\footnotesize}
    ]

    \node {Select}
        child { node {Byte(b'a')} }
        child { node {Join}
            child { node {Byte(b'a')} }
            child { node {Byte(b'b')} }
        };

    \end{tikzpicture}
    \caption{Grammar Tree Structure}
    \label{fig:grammar_tree}
\end{figure}
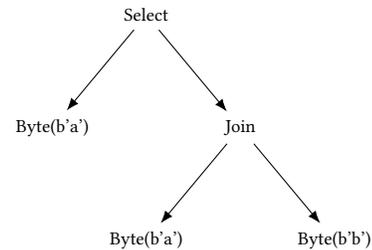

Guidance\footnote{\url{https://github.com/guidance-ai}} is a tool designed to optimize and enhance the process of generating text output with Large Language Models. It provides a flexible and efficient way to control and 'guide' the output of these models to achieve specific goals or adhere to desired formats. In our case, we use the formatting capabilities of guidance to produce output that adheres to the formatting rules of a given DSL.
Guidance internally defines a DSL implemented by developers from Microsoft for structured prompting of LLMs. Any prompting template consisting of a mix of unconstrained generation, function calls, constant strings, or grammar-constrained generation is transformed into a tree-like data structure, where each node represents different parts of the grammar. The core classes are \textit{Function} and its subclasses \textit{GrammarFunction}, which represents grammar rules, and \textit{RawFunction}, which is used to interleave native Python functions within a grammar. Grammar functions are either a join or a select, and regex expressions in the grammar are reduced to these two operations. Terminals are either individual bytes or byte ranges. Figure \ref{fig:grammar_tree} illustrates a simple grammar tree with the possible productions "a" or "ab".

The main idea for achieving a structured output is online parser-guided generation synchronizing a parser and scanner with an LLM to determine valid tokens at each step dynamically. In the main Loop, as seen in figure \autoref{fig:methodology}. When LLM generates text, it predicts the next Token in a sequence. For each possible token, a logit is generated, representing the confidence that this token is correct based on the training of the LLM. To those confidence values, a softmax function is applied to the confidence scores so they all add up to 1 and can be used as probabilities. Based on these probabilities, a multinomial distribution is calculated. Now, for each step, the tokens are tried one by one and checked by the guidance parser to see if they result in a valid partial parse.

The Parser is an Earley Parser constructed by Guidance based on the supplied grammar. An Earley parser efficiently processes context-free grammars in three phases. During prediction, it generates new states based on grammar rules for non-terminals. In the scanning phase, it matches and consumes terminal symbols in the input. Completion advances states when a rule ends, preparing for the next parsing steps. This method handles all possible paths, accommodating ambiguous and complex grammars effectively. Many other Frameworks only support subsets of CFGs since they are based on LR(1) or LALR(1) parsers. The Guidance parser enhances the standard Earley parser by introducing commit points, which force the parser to commit to specific parse paths, effectively pruning the search space and avoiding backtracking. No other alternatives are considered once a commit point is reached, ensuring that the parser adheres strictly to chosen paths.

The parser in the guidance framework employs several optimization strategies to enhance performance and efficiency by minimizing the frequency of calls to the LLM. In figure \autoref{fig:methodology}, we can also see that tries are used in two instances. 
A trie, or prefix tree, is a data structure that efficiently stores and retrieves keys, typically strings. Each node represents a common prefix shared by some keys, allowing fast lookups by a prefix. The tokens of the LLM are converted to a trie, making it possible to identify possible tokens and constraining the search space quickly. Another trie is constructed for the grammar, and by traversing it alongside generation, the forced prefixes and suffixes are added to the output without invoking the LLM.

This is in contrast to template-based approaches, which may force tokens that alter the attention distribution and potentially degrade the LLM output. Online parsers can maintain minimal invasiveness by checking tokens one by one. However, this method often incurs high inference overhead since they may need to check the entire model vocabulary at each step, as seen in 
\ref{table:time_parsed_values}.

\subsection{MontiCore based Modeling Languages}
As the chair for Software Engineering, we develop and maintain the language workbench MontiCore\footnote{\url{https://monticore.github.io/}} \cite{HKR21}. The language workbench is employed to generate a number of DSLs. For brevity, we will focus on two languages: one used to define requirements and specifications in simplified structured English (SEN) and another used to define UML class diagrams: CD4A.

\subsubsection{Structured English – A Controlled Natural
English (DSL) for Regulatory Comliance}
This DSL was primarily developed to standardize the definition of requirements and is based on the work of Konrad and Cheng \cite{konrad2005real}. The DSL can be used to write requirements and expressions in controlled simplified English \cite{fuchs2008attempto}, while still being able to be parsed and processed by tooling.
\vspace{1em}
\begin{lstlisting}[title=Generic Requirement in English]
After starting the engine, each time we 
pull the turn indicator lever up, the right 
indicator blinks within 500 ms.
\end{lstlisting}
\vspace{1em}

We can identify scopes and patterns in the provided Requirement:
\vspace{1em}
\begin{lstlisting}[title=Patterns found in Requirement]
After Q:Formula, if P:Formula holds, 
then in response S:Formula eventually holds 
time:TimeBound.
\end{lstlisting}
\vspace{1em}
Next, we can derive the SEN-Requirement from the natural English one:
\vspace{1em}
\begin{lstlisting}[title=Derived Structured English from Requirement]
After engine equals started, if 
turn_indicator_lever equals up holds, 
then in response right_indicator equals 
blinking eventually holds within 500 
Milliseconds.
\end{lstlisting}
\vspace{1em}



\subsubsection{Class Diagrams for Analysis (CD4A)}

The modeling language CD4A is based on UML class diagrams and closely implements all common features of class diagrams e.g., \textit{inheritance}, \textit{associations},  and \textit{enumerations} (see \cite{CD4A}). The syntax follows a Java-notation, making it easy for developers to adopt the language.
\autoref{lst:cd4a:example} depicts a simple class diagram in CD4A syntax. The diagram is titled '\texttt{LibraryDiagram}' and defines the four classes \texttt{Library}, \texttt{Member}, \texttt{Librarian}, and \texttt{Book}. In addition, the inheritance from Member to Librarian and an association from Library to Book is modeled.

\vspace*{1em}
\begin{lstlisting}[label=lst:cd4a:example,language=Java, caption={CD4A Class Diagram Defining Person, Student and Animal Class and their relations.}]
classdiagram LibraryDiagram {
  class Library {
    String name;
    String adress;
  }  
  class Member {
    String name;
    Long memberID;
    String contacInfo;
  }  
  class Librarian extends Member{}
  class Book {
    String title;
  }
  association [1]Library -> Book[*];
}
\end{lstlisting}

\section{Related Work}
Although open-source frameworks such as Guidance have been published in recent years, little research has been done on using LLMs with constrained decoding as a modeling tool.

Initial work was published by Wang et al. in \cite{wang2024grammar}. The presented approach uses grammar prompting to guide an LLM towards a constrained output. They demonstrate the viability of their approach with a selection of DSLs and LLMs.

\subsection{MBSE with generative AI}
Bader et al. use an FSL-based approach on the GPT-3.5-turbo-1106 LLM to produce textual UML Models in XML notation based on natural-language input. In this work, Bader shows that valid models can be created; however also points out some challenges of the LLM-based approach, such as limited context length and hallucination-related problems with the generated models \cite{baderfacilitating}.

Timperley et al. assess the usage of LLMs to generate model-based spacecraft system architectures \cite{timperley4823264assessment}. The approach relies on generating textual models for system architectures, requirements, and ontologies. The analysis concludes that LLMs can provide a high degree of assistance in modeling tasks in the early stages of spacecraft design. However, the modeling process still requires human supervision and can not be yet fully automated.
A similar conclusion is drawn by Busch et al. \cite{busch2023chatgpt}. In their approach, a Low-Code is developed using a visual modeling language. Similar to Timberly et al. full automation is not yet possible due to the uncertainty introduced by relying on an LLM to generate code.

\cite{austin2021program} explores the capabilities of current LLMs to create general-purpose code.

\subsection{SynCode} \label{sec:syncode}
SynCode \cite{ugare2024improving, githubGitHubUiucfocallabsyncode} is a framework for grammar-guided generation with large language models.
SynCode tries to address these limitations by using an offline-constructed lookup table called the DFA mask store. This table is based on the DFA of the language grammar terminals and is designed to retain only syntactically valid tokens during the generation process.

The core of SynCode's approach involves a two-step process during the LLM decoding stage. First, the partial output generated by the LLM is parsed to produce accept sequences and a remainder. Accept sequences represent valid terminal sequences that can follow the current partial output, while the remainder accounts for any unparsed or partially parsed tokens. Next, using the accept sequences and remainder, SynCode traverses the DFA states and retrieves masks from the DFA mask store. These masks filter out syntactically invalid tokens from the LLM's vocabulary, ensuring that only valid tokens are considered during each step of the generation process.

One of the greatest advantages of this approach is that the entire constraint infrastructure can be pre-computed. By accepting a longer initial setup time to generate the mask stores, the inference process becomes significantly faster, with only a minimal overhead of about 10\%, even for complex grammars. This makes it particularly well-suited for handling complex grammars, such as those found in MontiCore.


\section{Approach}
\begin{figure}[h]
  \centering
  \includegraphics[width=\linewidth]{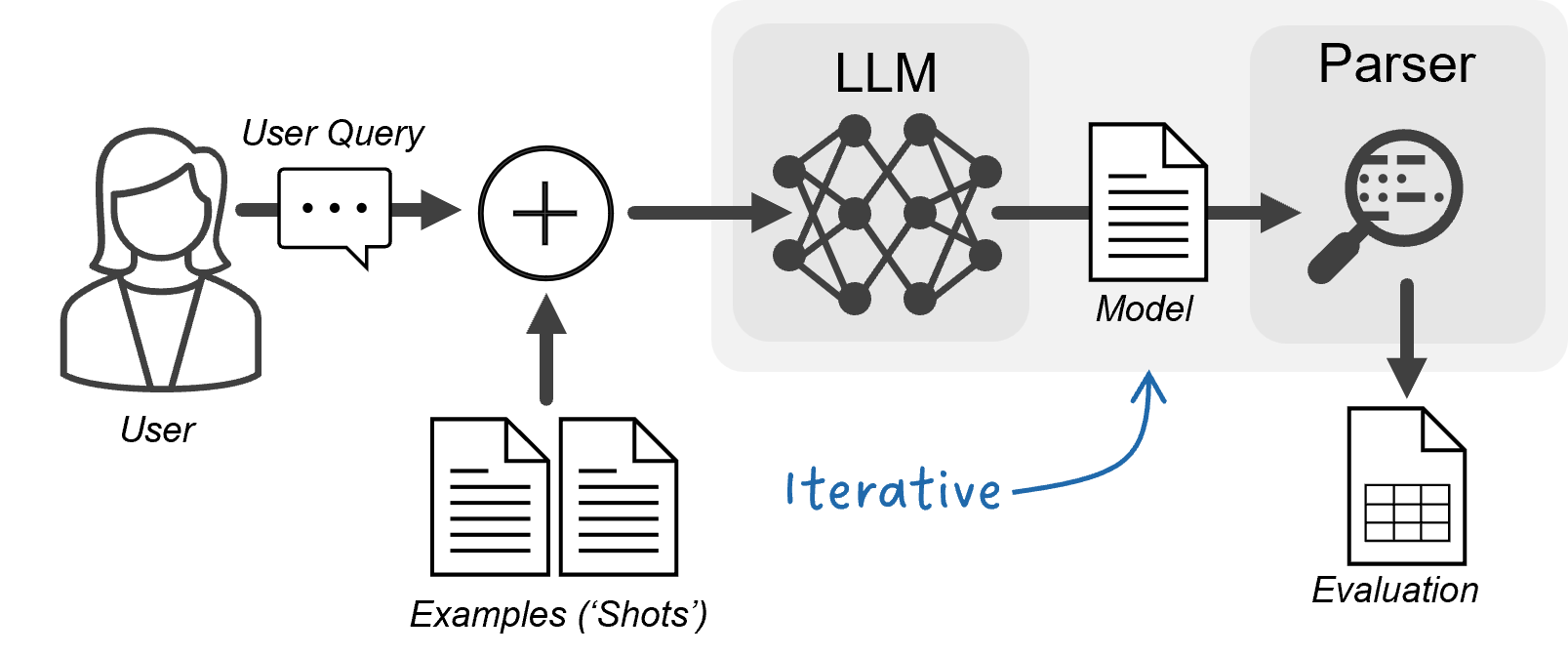}
  \caption{Evaluating the performance of a few-shot learning-based approach}
  \label{fig:fsl}
\end{figure}
Within this work, we focus on the syntactic correctness of the models produced by the LLM. We evaluate by comparing two approaches: one that only uses FSL and one that combines FSL with grammar masking. Even if a model is syntactically correct, there can still be semantic errors, e.g. the model could be an empty model, which might be syntactically valid, but does not satisfy the given modeling task. We were able to exclude this trivial error case in our tests, however, an in depth semantic analysis was not performed. Previous tests have shown that LLMs implement a large part of the requirements in the generated artifacts in the majority of cases \cite{netz2024natural}. The generated results did not indicate a deviation from the previous measurements on semantic accuracy.

\subsection{Using Few-Shot learning-based Modeling Method}

So far, FSL is one of the best prompting approaches to get an LLM to produce syntax in a predefined grammar. One drawback is that its performance heavily relies on the complexity of the grammar, the dependency on the LLM's familiarity with the concepts underlying the modeling task, and a good selection of examples to represent the rules of the grammar of the targeted DSL. FSL is limited to a set of examples to convey all syntactic relevant elements of a grammar \cite{wang2024grammar}, while also passing on 'best practices' for a modeling task in this language. As LLMs have a tendency to lose accuracy on increasingly larger prompts \cite{liu2024lost}, we have to choose the examples for each grammar, or even for each use case carefully. In our FSL approach, generic domain-independent examples for each grammar were selected, as our overreaching goal is the development of a domain-independent DSL-specific modeling approach, that is not optimized for a specific use case or domain. A higher accuracy is very likely, by narrowing down the approach to specific target domains and thus choosing corresponding examples from corresponding use cases.


Within this work, we compare our approach with the performance of the FSL approach developed in \cite{netz2024natural}. The approach is depicted in \autoref{fig:fsl}. A user informally defines a modeling task, that is extended with sample models of the target DSL. The extended prompt is provided to a LLM and the resulting model is checked by a parser. All models are provided with the same prompt-engineering and with the same set of molding tasks. The computation, with the exception of the OpenAI model, was run on the same hardware.

\begin{figure}[h]
  \centering
  \includegraphics[width=.8\linewidth]{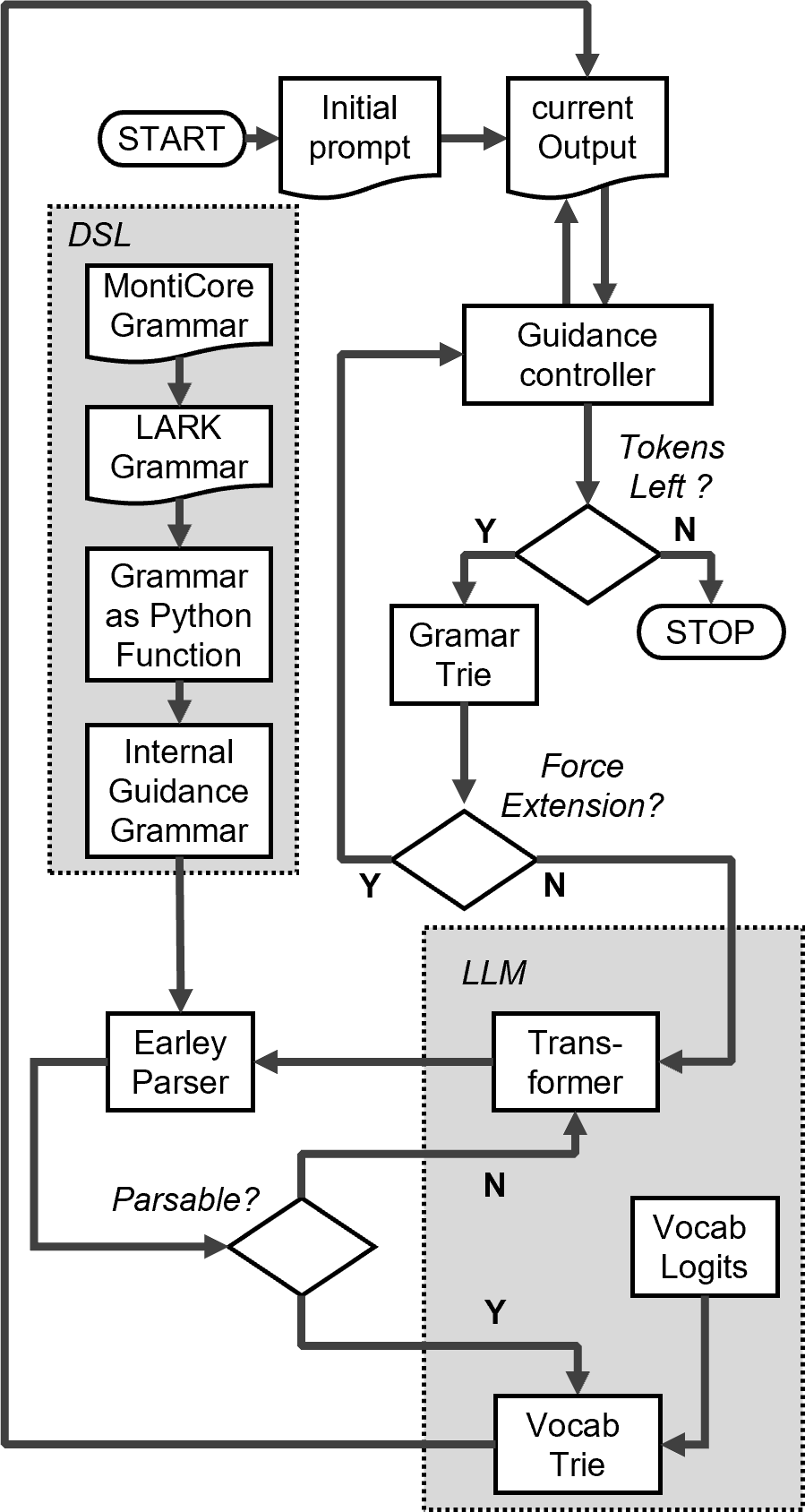}
  \caption{Combining the Guidance Framework with MontiCore to generate syntactically valid models.}
  \label{fig:methodology}
\end{figure}

\subsection{Using a Grammar Masking-based Modeling Method}
\begin{figure}
  \centering
  \includegraphics[width=\linewidth]{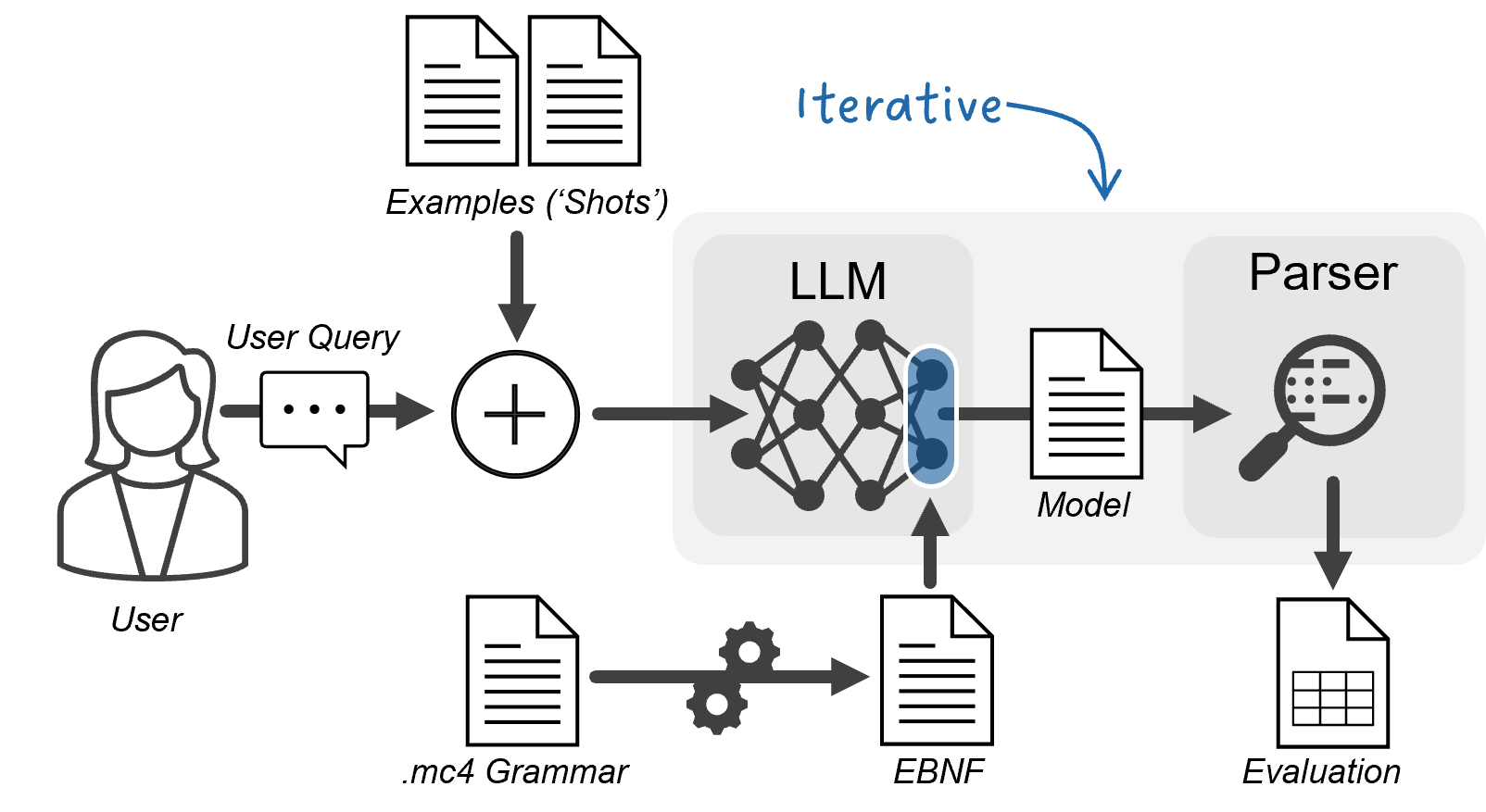}
  \caption{Evaluating the performance of a grammar-masking-based approach}
  \label{fig:guiadnceeval}
\end{figure}

In our constrained decoding approach we use Guidance as discussed in Section \ref{sec:guidance}. The constrained decoding approach is evaluated similarly as the FSL approach (cf. \autoref{fig:guiadnceeval}. The prompt containing the modeling task is supplemented with the same additional models, as in the previous approach. The grammar of the target DSL is transformed and provided through Guidance to the LLM. All generated models are parsed with a parser that is based on the same grammar. The pipeline involves transforming the MontiCore Grammar using a Visitor Pattern into a Lark Grammar \cite{larkparserWelcomeLarkx2019s} (cf. \autoref{lst:lark_grammar}), which is then integrated with Guidance. A detailed setup is shown in \autoref{fig:methodology}. The framework starts with an \textit{initial prompt}, that at the beginning is also the \textit{current prompt}. The framework is limited to a fixed set of tokens; if there are still tokens left, the system checks with the help of a \textit{grammar trie} if there is an unambiguous continuation for the current prompt (e.g., 'bool' has to be completed to 'boolean'). This is used as a shortcut to circumvent LLM usage. If this is not the case, the LLM is used to recommend tokens (\textit{transformer)}, which are passed to an \textit{earley parser}, which can identify invalid token suggestions. Valid tokens are passed on and added to the current output. The cycle starts again at the \textit{current output}. 

\begin{figure}
\begin{lstlisting}[
    caption={Example of Lark Grammar},
    label={lst:lark_grammar},
    breaklines
]
start: automaton
automaton: "automaton"  NAME  "{"  (state | transition)*  "}" 
state: "state"  NAME  ("<<"  ("initial" | "final")  ">>")*  ( ("{"  (state | transition)*  "}") | ";" ) 
transition: NAME  "-"  NAME  ">"  NAME  ";" 
NAME: /[a-zA-Z_$][a-zA-Z_0-9$]*/
%ignore WS
%import common.WS
\end{lstlisting}
\end{figure}

Currently, Guidance only uses greedy decoding, which picks the most probable allowed token. This reduces the effect of logit probability biasing, such as temperature, hence the same prompt will produce the same generated artifact. Thus to test the system, we need many distinct use cases.
Several modeling tasks from software engineering exams were selected as templates to synthesize further exam tasks. An LLM was commissioned to generate a list of 1000 domains (cf. \autoref{listing:domains}). 
\vspace{1em}
\begin{lstlisting}[caption={Except from synthesized domains. A complete list can be found at \cite{Netz_LLMs4MBSE_Synthetic-Artifacts_2024}}, label={listing:domains}]
 Automotive Systems,
 Hydraulic Press Control Systems,
 Healthcare Management Systems,
 E-commerce Platforms,
 Financial Trading Systems,
 Telecommunication Networks,
 Smart Home Automation,
[...]
\end{lstlisting}
\vspace{1em}
\begin{lstlisting}[caption={Except from synthesized use case .A complete list can be found at \cite{Netz_LLMs4MBSE_Synthetic-Artifacts_2024}}, label={listing:task}]
A voting system is being designed for a local 
election. The system should be able to handle 
multiple voting stations, each with its own 
set of voters. Each voter has a unique 
identifier and can cast one vote per election. 
The vote is recorded as a preference for a 
particular candidate.
[...]
\end{lstlisting}
\vspace{1em}
These were in turn given individually to the LLM in order to create new tasks using an FSL approach with the above-mentioned exam tasks. A set of 1000 exam tasks with similar specification levels was thus created (cf. \autoref{listing:task}). The task synthetization was executed by Llama 3 8B in a 4-bit quantization.


We then compare the artifacts that parse in constrained generation with those that parse in unconstrained generation for each task.

\section{Results}
To evaluate the presented approach several class diagrams and structured English models were generated, the parsing subset (4.225 CD4A models and 359 SEN models) can be found here: \cite{Netz_LLMs4MBSE_Synthetic-Artifacts_2024}.
The results (cf. \autoref{table:time_parsed_values}) indicate that the constrained generation method significantly increases the percentage of syntactically correct outputs from 46.52\% to 92.63\% (Llama 3). However, this improvement comes at the cost of increased generation time, with constrained generation taking an average of 74.09 seconds compared to 5.71 seconds for unconstrained generation. 
Similar results are observed for other LLMs. 36.57 \% of the models produced by Phi3 Mini in 4 Bit Quantization in an unconstrained mode are parsable, compared to 86.98 \% in the constrained mode. Gemma 7B in 4-Bit Quantization produces only 0.003 \% in an unconstrained mode, compared to 93.00 \% in the constrained mode. Mistral 7B in 3-Bit Quantization produces 20.99 \% parsable models compared to 92.37 \% parsable models in a constrained configuration. Quantization was chosen to accommodate the hardware constraints of the experimental setup. Models were limited to 8GB. All models took significantly more computation time when using constrained decoding. Using the Phi3 model in a constrained configuration took 34 times longer than using the Model in an unconstrained configuration. These increases could be combated by pre/computing the constraints, e.g. by using the Syncode approach (cf. \autoref{sec:syncode}) which is connected to further challenges. 

Using the same first 100 prompts, GPT-4o \cite{openai2024gpt4technicalreport} was also tested. We were unable to apply constrained decoding as it is a closed-source model. At the time of writing, GPT-4o is one of the most capable LLMs in several benchmarks \cite{artificialanalysisLeaderboardCompare}. Thus, it is expected that the model performs better (76 \% parsable models) in an unconstrained mode. However, most likely due to the communication overhead, the time needed to create a model is, on average, doubled in comparison to the locally running open-source models.

We also tested the Structured English DSL SEN in addition to the Class Diagram DSL CD4A. We observe the same patterns as in CD4A: In an unconstrained setting using Llama 3, 26.54\% of the produced models are parsable, whereas in a constrained configuration, 90.26\% are parsable. The same can be shown for the LLMs Phi3 and Mistral: significantly more produced models are parsable using constrained decoding, than using unconstrained generation. In comparison to the CD4A modeling task, GPT-4o does not perform significantly better than the locally running LLMs: with 24.63\% parsable models in an unconstrained configuration.

Not all runs could be completed in both cases (CD4A and SEN). Although 1000 prompts for CD4A and 123 promts for SEN were provided to the LLMs, a run was aborted in case a token limit was reached. Thus for example Llama3 8B 4-Bit only produced 991 models instead of 1000.

Constrained decoding does not currently achieve complete correctness because Monticore's grammar includes keywords not yet supported by our approach. For example, \texttt{enum (on, off, finished)} is interpreted as a function. In contrast, Monticore's implementation would not allow \texttt{enum} to be read as a function name, thereby guiding the generation incorrectly. Most of these differences were adapted by hand.

In \autoref{table:count_values}, we can see a peculiarity of constrained generation. While the overall number of language constructs used is similar, the number of compositions and associations is the same. This occurs because the model tries to extend tokens maximally. Due to the prompting and grammar constraints, both constructs, which have exactly 11 characters, are equally likely to be generated.

At this point, we would like to point out that the results do not show the best possible modeling capabilities of the individual models, as we have deliberately not optimized the few-shot learning prompting intensively. The results mainly show the performance gain with constant prompting with and without grammar prompting. The results with GPT-4o, which received the same prompts, serve as a comparison.


We encountered many unforeseen problems in achieving these results. First, EBNF grammars, which are supported by most of the currently available frameworks, are weak in expressing common features of interesting languages.

\begin{itemize}

\item \textbf{Whitespaces} MontiCore parsers are, in parts, whitespace-agnostic and ignore them, similar to most compilers such as the C compiler. However, whitespace is important for the attention distribution of the LLM when evaluating input and output. Therefore, grammars should be modified to enforce correct formatting, for example, by integrating them with a linter.

\item \textbf{Fuzzy Testing} Additionally, grammars are often only tested against human input instead of some extensive fuzzy testing, which usually results in no additional benefits. However, for an LLM, the grammar should be entirely correct. Often, in grammars that are common on the internet, some rules never trip a human up, but an LLM will make every error you allow it to.

\begin{figure}
    \begin{lstlisting}[language=Python, breaklines]
modifier: stereotype? ("public" | ... | "/" | ...)*
\end{lstlisting}

\vspace{1em}
Leading to this Code being parseable:
\vspace{1em}
\begin{lstlisting}[language=java, title=CD4A Code]
  class Frame  {
    Material material; // steel
    Wheel   ;
  }
\end{lstlisting}
\caption{CD4A Example}
\label{fig:grammar_error}
\end{figure}

\item \textbf{Grammar Mistakes} Instead of a modifier being only allowed once or not at all, the Kleene star allows \texttt{//} to be read as a modifier as shown in \autoref{fig:grammar_error}. By abusing incorrect grammar the LLM makes comments which are not allowed by using two modifiers.

\item \textbf{Endless repetitions and limited tokens}
A CD4A file can have arbitrarily many classes, and in fact, there can be arbitrarily many of most constructs. However, since our VRAM is limited, the LLM can only generate a finite amount of tokens. Therefore, grammar masking only guarantees correct artifacts if the generation stops before the token limit is reached. In some cases, the LLM gets stuck in endless repetitions, making it unlikely to terminate in a parsable state.
\end{itemize}

\begin{table}[h]
\centering
\caption{Mean Processing Time and Parsing Rates}
\label{table:time_parsed_values}
\begin{tabular}{lcc}
\toprule
 & \textbf{Unconstrained} & \textbf{Constrained} \\
 \midrule
\multicolumn{3}{c}{\textbf{CD4A (1000 examples)}} \\
\midrule
\textbf{Llama3 8B 4-Bit} & & \\
Time (s) & 5.71 & 74.10 \\
Parsed (\%) & 416/991 (41.97\%) & 918/991 (92.63\%)\\
\midrule
\textbf{Phi3 Mini 4-Bit} & & \\
Time (s) & 4.63 & 138.29 \\
Parsed (\%) & 357/976 (36.57\%) & 849/976 (86.98\%) \\
\midrule
\textbf{Gemma 7B 4-Bit} & & \\
Time (s) & 2.46 & 54.20 \\
Parsed (\%) & 2/658 (0.003\%) & 612/658 (93.00\%)\\
\midrule
\textbf{Mistral 7B 3-Bit} & & \\
Time (s) & 4.89 & 73.59 \\
Parsed (\%) & 190/905 (20.99\%) & 836/905 (92.37\%) \\
\midrule
\textbf{GPT-4o (100 examples)} & & \\
Time (s) & 8.77 & N/A \\
Parsed (\%) & 76/100 (76.00\%)& N/A \\
\midrule
\multicolumn{3}{c}{\textbf{SEN (123 examples)}} \\
\midrule
\textbf{Llama3 8B 4-Bit} & & \\
Time (s) & 1.04 & 5.23 \\
Parsed (\%) & 30/113 (26.54\%) & 102/113 (90.26\%) \\
\midrule
\textbf{Phi-3-Mini 4-Bit} & & \\
Time (s) & 1.13 & 10.12 \\
Parsed (\%) & 4/122 (3.27\%)& 105/122 (86.06\%)\\
\midrule
\textbf{Mistral 7B 3-Bit} & & \\
Time (s) & 1.22 & 5.39 \\
Parsed (\%) & 11/96 (11.45\%) & 85/96 (88.54\%) \\
\midrule
\textbf{GPT-4o} & & \\
Time (s) & 1.84 & N/A \\
Parsed (\%) & 17/69 (24.63\%) & N/A \\
\bottomrule
\end{tabular}
\begin{tablenotes}
\item Numbers include only generations without out-of-token errors.
\end{tablenotes}
\end{table}

\begin{table}[h]
\centering
\caption{Mean Values of Syntactic Elements (Llama3 CD4A)}
\label{table:count_values}
\begin{tabular}{lcc}
\toprule
 & \textbf{Unconstrained} & \textbf{Constrained} \\
\midrule
Composition Count & 5.669021 & 5.809284 \\
Association Count & 0.21998 & 5.809284 \\
Class Count & 7.849647 & 8.095863 \\
\bottomrule
\end{tabular}
\end{table}

\section{Discussion}
The results shown in \autoref{table:time_parsed_values} and \autoref{table:count_values} show very promising results, in the following we discuss aspects such as limitations and generalizability of the approach. 

\subsection{Applicability to other grammars}

The approach presented in this work is based on MontiCore Grammars, which are transformed into LARK grammars. Hence this approach can be applied to any MontiCore Grammar. MontiCore provides infrastructure that permits the developers to define context conditions (CoCos) \cite{BGJ+23}. These CoCos are rules that check the well-formedness of models. These context conditions are crucial for ensuring that models adhere to the specified rules and constraints of the language. DSLs with fewer CoCos are transferable to this method with less effort, as this approach only impacts the adherence to the grammar and not the CoCos. 
In addition, grammar masking is expected to yield fewer improvements on DSLs already encountered in pretraining, such as PlantUML \cite{singh2018optimizing}, or SQL, as these languages should already perform well. LLMs that are already trained on a specific DSL will be more likely to produce syntactically correct models.

This approach was developed with MontiCore grammars in mind. However, the approach can be applied to any grammar that is transformable into a LARK grammar (cf. \autoref{fig:methodology}, \autoref{fig:guiadnceeval}).

\subsection{Impact Grammar Masking on Model Semantics}
The approach presented in this paper relies primarily on filtering out syntactically invalid tokens. Although we did not notice any changes in the semantics of the resulting models, we cannot rule out the possibility that relevant content is excluded from the model or modified so that the semantics of the model are changed. A closer look at random samples, in which both approaches resulted in syntactically correct models, revealed that models generated with the constrained approach do not systematically contain fewer elements than those generated with the unconstrained approach. \autoref{table:count_values} implies the opposite. The differences observed were mainly in the formatting and naming of elements.

 Models generated with contained decoding might be less detailed, as this does not necessarily appear in quantitative analysis (e.g. counting classes and attributes). A further in-depth analysis of the content would be necessary. 

\subsection{Choosing Between Constrained and Unconstrained Generation}
As we could show in the case of the DSL CD4A, FSL alone can suffice to create an approach that is very likely to yield a syntactically correct model (e.g. by using a sufisticated LLM such as GPT-4o cf. \autoref{table:time_parsed_values}). This approach is not limited to the DSLs presented in this paper and can be applied to further modeling languages with sufficient prompt engineering. Nevertheless, this approach requires experts in the field of generative AI (e.g. a Prompt Engineer) and experts in the specific modeling language to provide specific and ideal representative examples of the targeted language. Hence, an FSL-only-based approach would be unsuitable for developers unfamiliar with the targeted DSL. In contrast, the grammar-masking-based approach is less reliant on good prompting and can be derived from a given MontiCore grammar, thus only needing the grammar file and a few samples of the targeted modeling language to operate. In addition, the grammar masking approach presented in this work enables smaller less performant models to serve as modeling tools. Smaller models such as the Llama 3 8B in a 4-bit quantization can be executed on hardware that is available for the end user (e.g. NVIDIA GeForce RTX 3070), making this approach independent from external server clusters such as the ones needed for an OpenAI based approach.

\subsection{Limitations}
One of the key limitations of this approach is its missing support for context conditions. As context conditions often need the entire model to be applied, they can not be used in a filtering capacity during model creation. Thus, they have to be applied in a succeeding step after a model for a specific DSL has been created. Grammar masking reduces the number of models that do not adhere to the provided grammar, thus also increasing the overall number of correct models that potentially comply with the context conditions, as any syntactically invalid models are filtered out at a very early stage.
Another limitation is the high degree of specialization in one grammar. This approach limits the LLM to only producing models for one specific grammar. A second setup and a delegator are needed if the framework is meant to switch between DSLs. Switching out the prompting alone will not suffice.

As mentioned above, grammars might need to be adapted and refined to operate with this approach, as LLMs tend to find loopholes in 'incompletely' defined grammars. This requires the developer who sets up the framework to have some experience in language design.

\section{Conclusion}

We were able to show that frameworks that enable constrained decoding enable smaller, less performant LLMs to produce syntactically correct models at a reasonable rate. Our experiments show, that grammar masking can significantly increase the chance of an LLM-based approach to produce valid models for a given DSL. We could demonstrate this improvement for two DSLs. Within this paper, we only address syntactic validation of the produced models. Further analysis has to be performed to systematically evaluate if there are systematic differences in the semantics of unconstrained and constrained models. In addition, an extensive difference in computation time between unconstrained and constrained generation was measured. As a result, we recommend that the method described in this work should only be used if a satisfactory outcome cannot be achieved using conventional prompt engineering methods. Improvements in frameworks, such as precomputations (Syncode) and runtime optimations, could soon reduce the gap in computation time. 
\bibliographystyle{ACM-Reference-Format}
\bibliography{main}
\end{document}